\documentclass[runningheads]{llncs}
\usepackage[T1]{fontenc}
\usepackage{graphicx}
\usepackage{booktabs}
\usepackage[misc]{ifsym}
\newcommand{\corr}{(\Letter)}
\usepackage{mwe}
\usepackage{amsmath}
\usepackage{graphicx}
\usepackage{makecell}
\usepackage{subcaption}
\usepackage{multirow}
\usepackage{amssymb}
\usepackage{cite}

\begin{document}

\title{Arch-VQ: Discrete Architecture Representation Learning with Autoregressive Priors}

\titlerunning{Discrete Architecture Representation Learning}

\author{Deshani Geethika Poddenige\inst{1} \and Sachith Seneviratne\inst{1} Asela Hevapathige\inst{1} \and Damith Senanayake\inst{1} \and Mahesan Niranjan\inst{2} \and PN Suganthan\inst{3} \and Saman Halgamuge\inst{1} \corr}

\authorrunning{Deshani Geethika Poddenige et al.}

\institute{AI, Optimization and Pattern Recognition Research Group, Dept. of Mechanical Eng., University of Melbourne, Australia 
\and
School of Electronics and Computer Science, University of Southampton, UK
\and
KINDI Computing Research Center, College of Engineering, Qatar University, Doha, Qatar \\
\email{dpoddenige@student.unimelb.edu.au, \{sachith.seneviratne, asela.hevapathige,  damith.senanayake\}@unimelb.edu.au, \\
mn@ecs.soton.ac.uk, p.n.suganthan@qu.edu.qa, saman@unimelb.edu.au}}

\maketitle              

\begin{abstract}
 Existing neural architecture representation learning methods focus on continuous representation learning, typically using Variational Autoencoders (VAEs) to map discrete architectures onto a continuous Gaussian distribution. However, sampling from these spaces often leads to a high percentage of invalid or duplicate neural architectures, likely due to the unnatural mapping of inherently discrete architectural space onto a continuous space. In this work, we revisit architecture representation learning from a fundamentally discrete perspective. We propose Arch-VQ, a framework that learns a discrete latent space of neural architectures using a Vector-Quantized Variational Autoencoder (VQ-VAE), and models the latent prior with an autoregressive transformer. This formulation yields discrete architecture representations that are better aligned with the underlying search space while decoupling representation learning from prior modeling. Across NASBench-101, NASBench-201, and DARTS search spaces, Arch-VQ improves the quality of generated architectures, increasing the rate of valid and unique generations by 22\%, 26\%, and 135\%, respectively, over state-of-the-art baselines. We further show that modeling discrete embeddings autoregressively enhances downstream neural predictor performance, establishing the practical utility of this discrete formulation.
\end{abstract}

\section{Introduction}

\begin{figure}[h!]
    \centering
    \includegraphics[width=1\textwidth]{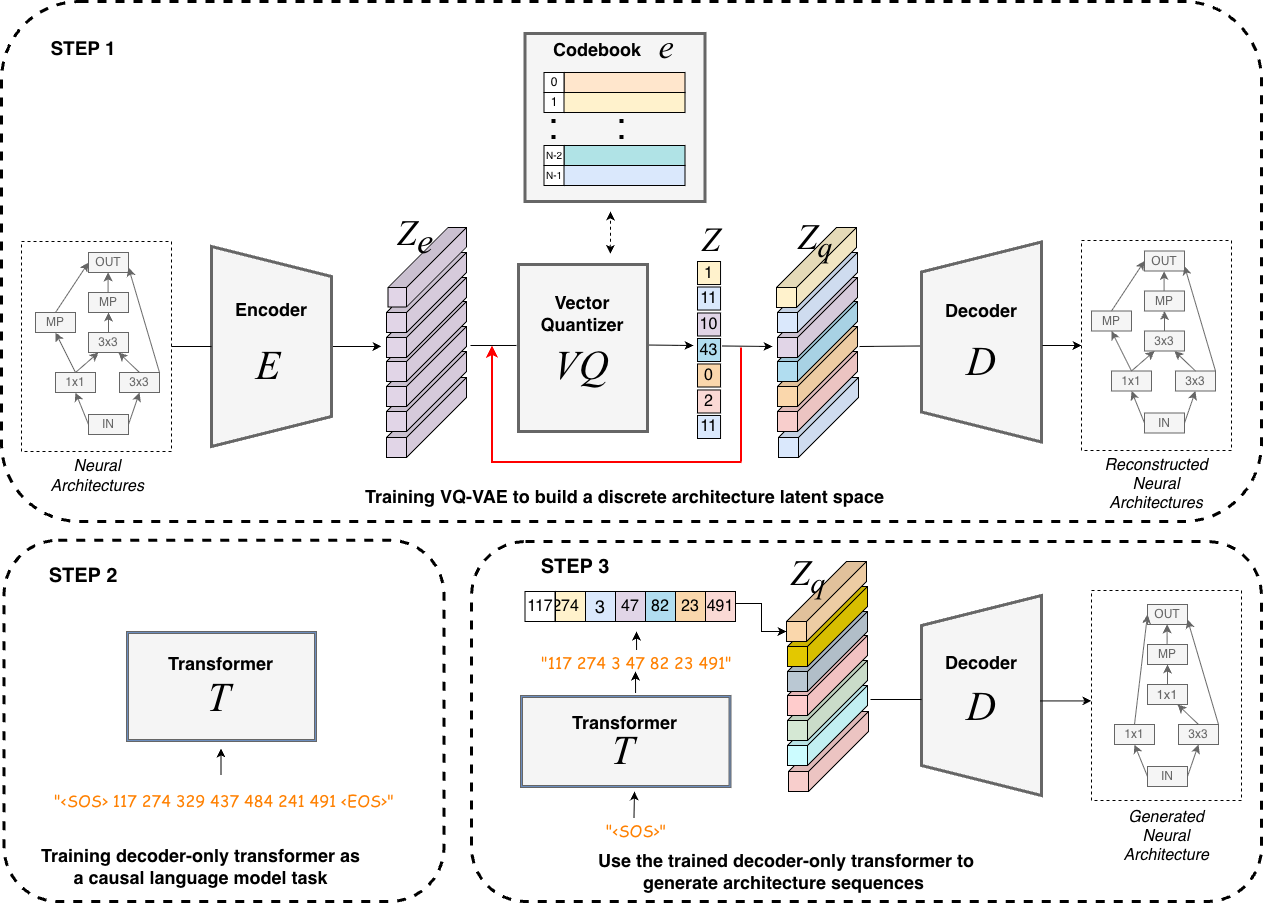}
    \caption{Illustration of the proposed \textit{Arch-VQ} framework. (1) \textbf{Step 1:} A VQ-VAE is trained to learn a discrete latent space of neural architectures. Each architecture, represented by its adjacency matrix and one-hot node features, is passed through the encoder $E$ to obtain the latent representation $Z_e$. The vector quantization module then maps $Z_e$ to its nearest codebook entry, producing the discrete indices $Z$ and quantized latent vector $Z_q$. The decoder $D$ reconstructs $Z_q$ back into the original architecture representation. (2) \textbf{Step 2:} The discrete index sequence $Z$ is augmented with start and end tokens and used to train a transformer with a next-token prediction objective. (3) \textbf{Step 3:} The trained transformer generates new architecture sequences autoregressively, beginning from the start token.}
    \label{fig:pipeline}
\end{figure}

Unsupervised representation learning has driven transformative paradigm shifts across domains such as computer vision \cite{van2016conditional,he2020momentum}, language representations \cite{mikolov2013distributed,devlin2019bert,OpenaiImprovingPre-Training}, and audio processing \cite{jansen2018unsupervised,ayilo2025diffusion}. Yet its potential remains relatively underexplored in the neural architecture domain. Existing work on neural architecture representation learning has emerged in the context of Neural Architecture Search (NAS), where early methods operated directly in the original encoding space of neural architectures, typically represented using adjacency matrix and a list of operations \cite{zoph2017neural,yang2023hotnas,real2017large,real2019regularized}. However, the lack of explicit structural organization and poor scalability motivated a shift toward learning continuous architecture embeddings and conducting search in a latent space. A common strategy in this line of work is to learn representations jointly with optimization in a supervised manner \cite{luo2018neural,liu2018darts}, prioritizing the NAS objective rather than modeling the intrinsic distribution of architectures. Existing unsupervised representation learning approaches are most notably based on Variational Autoencoders (VAE) \cite{zhang2019d,yan2020does,lukasik2021smooth}.

Despite their empirical utility, continuous latent models like VAEs impose an implicit assumption: that the architecture distribution can be adequately approximated by a continuous Gaussian prior. Neural architectures, however, are discrete combinatorial objects defined by graph topology and categorical operations. Due to this fundamental mismatch, sampling from a continuous latent space and decoding back to discrete graphs often produces a high percentage of invalid \cite{yan2020does} or duplicate samples \cite{lukasik2021smooth,lukasik2022learning,zhang2019d}. As a result, large numbers of samples must be generated to obtain a decent proportion of valid and distinct architectures, increasing computational cost and limiting the latent space reliability. This mismatch points to a fundamental gap in the literature: no existing unsupervised method has explored a representation space that is natively discrete and aligned with the combinatorial structure of neural architectures.

Discrete representation learning has proven effective in other discrete modalities \cite{van2017neural,esser2021taming,tjandra2019vqvae,yan2021videogpt}, where Vector-Quantized Variational Autoencoders (VQ-VAEs) learn codebook-based latent embeddings and decouple representation learning from prior modeling. VQ-VAE offers several key advantages over VAEs, such as: (i) they construct a discrete latent space by mapping each input to a set of discrete latent codes; and (ii) instead of relying on a fixed Gaussian prior, they learn the prior, enabling flexibility to pair with sophisticated autoregressive models for prior modeling.

In this paper, we extend this paradigm to neural architectures and introduce Arch-VQ, a framework for unsupervised discrete representation learning of architecture graphs. To the best of our knowledge, Arch-VQ is the first framework to learn a discrete latent space of neural architectures via VQ-VAE, and the first to model the resulting discrete codes with an autoregressive transformer prior — two design choices that are individually motivated and jointly critical. First, we construct a discrete latent space using a VQ-VAE built upon graph encoders, mapping each architecture to a sequence of codebook indices. Second, we learn a prior over these discrete sequences using a decoder-only transformer trained with causal language modeling. This decoupling of representation learning from prior modeling is a key architectural distinction from prior VAE-based methods, which conflate both objectives within a single Gaussian framework. 

We evaluate Arch-VQ on NASBench-101 \cite{ying2019bench}, NASBench-201 \cite{dongbench}, and DARTS \cite{liu2018darts} search spaces against unsupervised VAE-based and unconditional generative baselines. Across all benchmarks, Arch-VQ significantly improves architecture generation quality, increasing the proportion of valid and unique samples by 22\%, 26\%, and 135\%, respectively, compared to state-of-the-art methods, with the largest gains observed on the more complex DARTS search space. Moreover, by varying the sampling temperature, Arch-VQ provides a controllable balance between fidelity and exploration, yielding 20\% and 135\% improvements in valid and novel generations on NASBench-201 and DARTS, respectively. We further show that the learned discrete sequential representation improves downstream neural performance prediction. Our contributions are summarized as follows:
\begin{itemize}
\item We propose Arch-VQ, a novel framework for unsupervised discrete representation learning of neural architectures using VQ-VAE — the first method to apply discrete latent modeling to the neural architecture domain.
\item We introduce autoregressive prior modeling over discrete architecture codes, enabling efficient architecture generation.  This combination of VQ-VAE and autoregressive prior is a new formulation that no prior NAS representation learning work has explored.
\item We empirically demonstrate improvements in the quality of latent space over state-of-the-art baselines across multiple neural architecture benchmarks,  with particularly large gains in the DARTS search space.
\item We show that modeling architectures as discrete sequences yields better neural performance prediction, demonstrating that the representational benefits of discretization extend beyond generation to downstream tasks.
\end{itemize}

\section{Related Work}

\subsubsection{Neural architecture representation learning}
Neural architecture representation learning has mainly developed within the NAS literature, where most methods rely on performance supervision, which shapes the representation to a specific optimization objective \cite{luo2018neural}. Our focus is on unsupervised architecture representation learning, where the goal is to model the structure of architecture spaces without performance labels. Existing unsupervised methods, including Arch2vec, D-VAE, and SVGe, typically adopt VAE-style continuous latent spaces for discrete architecture graphs \cite{yan2020does,zhang2019d,lukasik2021smooth}. This design introduces a mismatch between continuous latent priors and discrete architectural structures, which can result in invalid or duplicate decoded samples. This limitation motivates studying discrete latent representations for neural architectures.

\subsubsection{Neural architecture generation}
Recent NAS research has also explored \emph{neural architecture generation}, where architectures are sampled from a learned generator. These methods can be broadly divided into \emph{unconditional} generators, which model architecture distributions without performance labels, and \emph{performance-guided} generators, which incorporate feedback such as accuracy or latency. Since our focus is on unsupervised structure learning, we compare only with methods that include an unconditional generative stage. These include AG-Net \cite{lukasik2022learning}, MetaD2A \cite{lee2021rapid}, and DiffusionNAG \cite{an2024diffusionnag}. These methods are summarized in Table~\ref{tab:nas_related_work}.

\begin{table}[b]
\centering
\small
\setlength{\tabcolsep}{4pt}
\renewcommand{\arraystretch}{1.2}
\resizebox{\columnwidth}{!}{%
\begin{tabular}{lcccccc}
\hline
Method & Unsupervised & Hybrid & Embedding & Generative & RL/BO & Predictor-guided \\
\hline
Arch2Vec \cite{yan2020does}           & \checkmark  &    $\times$     & \checkmark &        $\times$    & \checkmark & $\times$ \\
SVGe \cite{lukasik2021smooth}         & \checkmark  &   $\times$         & \checkmark &       $\times$     &  \checkmark & $\times$\\
AG-Net \cite{lukasik2022learning}  &    $\times$         & \checkmark &        $\times$    & \checkmark &  $\times$  &\checkmark \\
MetaD2A \cite{lee2021rapid}         &    $\times$         & \checkmark &      $\times$      & \checkmark &  $\times$ & \checkmark \\
DiffusionNAG \cite{an2024diffusionnag} &     $\times$     & \checkmark &      $\times$     & \checkmark & $\times$ & \checkmark \\
\hline
\end{tabular}%
}
\caption{Categorization of methods (ticks indicate membership).}
\label{tab:nas_related_work}
\end{table}

\subsubsection{Discrete representation learning with VQ-VAE}
VQ-VAE \cite{van2017neural} has been successfully adopted for discrete and high-dimensional signals such as images, audio, and video, often paired with an autoregressive prior over discrete codes \cite{van2017neural,esser2021taming,tjandra2019vqvae,yan2021videogpt}. Neural architectures are also discrete combinatorial objects (graph structure with categorical operations), which makes discrete latent modeling a natural fit. However, to the best of our knowledge, the potential of using VQ-VAE for neural architecture representation learning is not explored in the literature.

\section{Arch-VQ Architecture}
\label{headings}
This section outlines our method; \textit{Arch-VQ} (See Figure \ref{fig:pipeline}). First, we build a discrete latent space of neural architectures using a VQ-VAE. We adopt the Variational Graph Isomorphism Autoencoder from arch2vec \cite{yan2020does}, integrating a Vector Quantizer \cite{van2017neural} to effectively discretize the latent space. We then train a decoder-only transformer for causal language modeling using our discrete neural architecture sequences.

\subsection{Discrete Architecture Autoencoding (VQ-VAE)}

\subsubsection{Preliminaries}
We consider cell-based architectures as in NAS-Bench-101 \cite{ying2019bench}, where each cell is a directed acyclic graph \(G=(V,E)\) with \(N\) nodes and \(E\) edges. Node operations are chosen from \(K\) predefined types. We represent each cell by an adjacency matrix \(\mathbf{A}\in\mathbb{R}^{N\times N}\) and a one-hot operation matrix \(\mathbf{X}\in\mathbb{R}^{N\times K}\).

\subsubsection{Encoder}
The encoder consists of Graph Isomorphism Network (GIN)\cite{xu2018how} layers to produce an encoded vector $\mathbf{Z_e}(\mathbf{\tilde{A}}, \mathbf{X})$ of an architecture ($\mathbf{A}, \mathbf{X}$). Similar to arch2vec, first we allow bi-directional information flow by converting the original directed graphs to undirected graphs, augmenting the adjacency matrix A to $\mathbf{\tilde{A}=A+A^T}$. L-layer GIN is used to get the node embedding matrix $\mathbf{H}$:

\begin{equation}
\mathbf{H}^{(k)} = \mathbf{MLP}^{(k)} \left((1 + \epsilon^{(k)})\cdot \mathbf{H}^{(k-1)} + \mathbf{\tilde{A}}\mathbf{H}^{(k-1)}\right)
\label{eq:eq:1}
\end{equation}

where k = 1, 2,\ldots,L, the initial value of $\mathbf{H}$, $\mathbf{H}^{(0)} = \mathbf{X}$, $\epsilon$ is a trainable bias, and $\mathbf{MLP}$ is a multi-layer perception where each layer consists of a linear-batchnorm-ReLU triplet. The encoder output $\mathbf{Z}_e(\mathbf{\tilde{A}}, \mathbf{X})$ is obtained by feeding the final node embedding matrix $\mathbf{H}^{(L)}$ to a fully connected layer. Unlike in arch2vec, we don't obtain the mean and the variance of a posterior approximation, as our posterior is categorical and defined by the vector-quantizer component. 

\subsubsection{Vector-quantizer}
We incorporate a vector-quantizer {\cite{van2017neural}} as a discretization bottleneck to our framework. It employs a learnable codebook, which acts as a lookup table defining the discrete latent embedding space $e \in \mathbb{R}^{K\times D}$. Here $K$ is the number of vectors in the discrete latent space (i.e. $K$-way categorical), and $D$ is the dimensionality of each embedding vector $e_j \in \mathbb{R}^D, j=1,2,...,K$. The discrete latent variable $\mathbf{Z}$ corresponding to $\mathbf{Z}_e(\mathbf{\tilde{A}}, \mathbf{X})$ is calculated by nearest neighbourhood lookup using the codebook vectors $e$ as described in equation \ref{eq:2}. The input to the decoder is the corresponding embedding vector $e_k$ as shown in equation \eqref{eq:3}. The posterior categorical distribution $q(\mathbf{Z}|\mathbf{\tilde{A}}, \mathbf{X})$ probabilities can be described as one-hot encoding as follows:

\begin{equation}
    q(\mathbf{Z}=k|\mathbf{\tilde{A}}, {\mathbf{X}}) = 
    \begin{cases}
        1  & k = \text{argmin}_j||\mathbf{Z}_e(\mathbf{\tilde{A}}, \mathbf{X})-\mathbf{e}_j||_2, \\ 0 & \text{otherwise}
    \end{cases}\label{eq:2}
\end{equation}

Equation \eqref{eq:2} outputs a set of nearest indices $\mathbf{Z}$ of the $\mathbf{Z}_e$. Then these indices are mapped into the corresponding embedding vectors $\mathbf{Z}_q$ as given in the following equation:
\begin{equation}
    \mathbf{Z}_q(\mathbf{\tilde{A}}, {\mathbf{X}}) = \mathbf{e}_k, \text{    where   } k = \text{argmin}_j||\mathbf{Z}_e(\mathbf{\tilde{A}}, \mathbf{X}) - \mathbf{e}_j||_2 
    \label{eq:3}
\end{equation}

\subsubsection{Decoder}
The decoder uses the input $\mathbf{Z}_q$ from the latent space and reconstructs the adjacency matrix $\mathbf{\hat{A}}$ and operations $\mathbf{\hat{X}}$ similar to the original inputs of $\mathbf{\tilde{{A}}}$ and $\mathbf{X}$ respectively.

\begin{equation}
p(\mathbf{\hat{A}} \mid \mathbf{Z}_q)
= \prod_{i=1}^{N} \prod_{j=1}^{N}
P(\hat{A}_{ij} \mid \mathbf{z}_i, \mathbf{z}_j), \text{with } 
P(\hat{A}_{ij}=1 \mid \mathbf{z}_i, \mathbf{z}_j) = \sigma\!\left(\mathbf{z}_i^{\top} \mathbf{z}_j\right)
\label{eq:4}
\end{equation}

\begin{equation}
p(\mathbf{\hat{X}} = [k_1, \ldots, k_N]^{\top} \mid \mathbf{Z}_q)
= \prod_{i=1}^{N} P(\hat{X}_i = k_i \mid \mathbf{z}_i),
= \prod_{i=1}^{N}
\text{softmax}\!\left(\mathbf{W}_o \mathbf{Z}_q + \mathbf{b}_o\right)_{i,k_i}
\label{eq:5}
\end{equation}

where $\sigma(\cdot)$ is the sigmoid activation function, and softmax(·) is the softmax activation function applied row-wise. $k_n \in {1, 2, ...,K}$ denotes the operation chosen from the predefined set of $K$ operations at the $\text{n}^{th}$ node. $\mathbf{W}_o$ and $\mathbf{b}_o$ are learnable weights and biases of the decoder.

\subsection{Training Objective of VQ-VAE}

Since the quantization process is non-differentiable, we use a straight-through estimator for backpropagation \cite{van2017neural}. We adopt VQ-VAE with EMA codebook updates, which stabilize training by updating embeddings from assigned encoder outputs rather than by gradient descent. For embedding $i$, the smoothed count $N_i$ and average $m_i$ are updated as

\begin{equation}
    N_i^{(t)} = \gamma N_i^{(t-1)} + (1 - \gamma) n_i^{(t)}, \\
    m_i^{(t)} = \gamma m_i^{(t-1)} + (1 - \gamma) \sum_{j} \mathbf{Z}_{e,j}^{(t)}
\end{equation}

where $\gamma$ is the decay parameter and $n_i^{(t)}$ denotes the number of encoder outputs assigned to embedding $i$ at step $t$. The embeddings are then normalized as:
\begin{equation}
\mathbf{e}_i^{(t)} = \frac{m_i^{(t)}}{N_i^{(t)}}
\end{equation}

The VQ-VAE objective combines reconstruction and commitment losses defined in equation \eqref{eq:6}:

\begin{equation}
\begin{gathered}
\mathcal{L}
= \log p(\mathbf{\hat{X}}, \mathbf{\hat{A}} \mid \mathbf{Z}_q)  
+ \beta \left\lVert
\mathbf{Z}_e(\mathbf{X}, \mathbf{\tilde{A}}) - \operatorname{sg}\!\left[\mathbf{e}\right]
\right\rVert_2^{2}
\end{gathered}
\label{eq:6}
\end{equation}

where $\operatorname{sg}$ denotes stop-gradient. The decoder is trained with the reconstruction term, the encoder with both terms, and the codebook with EMA updates.

\subsection{Autoregressive Prior Over Code Sequences (Decoder-only Transformer)}
\label{Sec: train-transformer}
Once the latent space is discretised using VQ-VAE, we can represent each neural architecture using the codebook indices of their encodings. More specifically, the quantized encoding of an architecture $(\mathbf{X}, \mathbf{\tilde{A}})$ is given by $\mathbf{Z}_q \in \mathbb{R}^{K\times D}$ and is obtained by mapping a sequence of closest codebook indices $s \in \{0,1,\dots,|Z|-1\}$  to their respective vectors. 

\begin{equation}
    s_{i} = k \text{ such that } {(\mathbf{z}_q)}_{i} = z_k
\end{equation}

We formulate an architecture as a numerical sequence $s$ and then convert $s$ into a sentence-like format by appending start <SOS> and end <EOS> tokens. This dataset is subsequently used to train a decoder-only transformer model for causal language modeling, training it to predict the next token in a sequence using an auto-regressive approach. Finally, we use our transformer to generate architecture sequences by prompting it with the start token <SOS>. 

\subsection{Complexity Analysis}
Let $N$, $E$, $L$, $D$, $K$, $S{=}N$, $d$, and $T$ denote the number of nodes, 
edges, GIN layers, embedding dimension, codebook size, sequence length, 
transformer hidden dimension, and transformer layers, respectively. 
The per-sample complexity of Arch-VQ is 
$\mathcal{O}\!\left(L(N+E)D + SKD + N^2D + TS^2d\right)$, 
corresponding to the GIN encoder, vector quantizer, decoder, and transformer 
prior respectively. Within the cell-based search spaces evaluated in this work, 
$N$, $E$, and $S$ are small constants, and training complexity scales as 
$\mathcal{O}(|\mathcal{D}|)$ since each architecture is processed independently 
with no inter-sample dependencies.

\section{Experimental Design}

\subsection{Datasets}
We evaluate \textit{Arch-VQ} on three standard neural architecture search spaces in NAS literature. For all datasets, we use a 90\% train and 10\% validation split to train the VQ-VAE and the decoder-only transformer.

\begin{itemize}
    \item \textbf{NAS-Bench-101}~\cite{ying2019bench}: A tabular cell-based benchmark with 423k unique CIFAR-10 evaluated architectures. Cells are constrained to $|V|\leq 7$ nodes (including input/output) and $|E|\leq 9$ edges, with each intermediate node being an operation from $O=\{1\times1\ \text{conv},\ 3\times3\ \text{conv},\ 3\times3\ \text{max-pool}\}$.
    \item \textbf{NAS-Bench-201}~\cite{dongbench}: Contains 15,625 unique cell architectures (4 nodes, 5 operations) evaluated on CIFAR-10, CIFAR-100, and ImageNet-16-120. Nodes represent feature maps and edges are labeled with $O=\{1\times1\ \text{conv},\ 3\times3\ \text{conv},\ 3\times3\ \text{avg-pool},\ \text{skip},\ \text{zero}\}$.
    \item \textbf{DARTS}~\cite{liu2018darts}: A large-scale search space defining convolution and reduction cells, each with six nodes forming a DAG where the first two nodes come from the previous two cells and the remaining nodes have two incoming edges. Following~\cite{yan2020does}, we randomly sample 600k unique architectures.
\end{itemize}

\subsection{Evaluation Metrics}

We evaluate the quality of the learned discrete representation space of Arch-VQ using four established metrics~\cite{zhang2019d}:

\begin{itemize}
    \item \textbf{Reconstruction Accuracy}: Percentage of architectures in the validation set correctly reconstructed by the decoder.
    \item \textbf{Validity}: Percentage of valid architectures among those generated from the latent space.
    \item \textbf{Uniqueness}: Percentage of unique architectures among the valid ones generated.
    \item \textbf{Novelty}: Percentage of architectures among the valid ones that were not seen during training.
\end{itemize}

\noindent Since \textit{Uniqueness} and \textit{Novelty} are conditioned on valid samples, we also report \textit{Absolute Uniqueness} and \textit{Absolute Novelty}, defined as $(\text{\textit{Validity}} \times \text{\textit{Uniqueness}})/100$ and $(\text{\textit{Validity}} \times \text{\textit{Novelty}})/100$ respectively, to reflect their overall proportion among all generated architectures.

\subsection{Model Hyperparameters and Baselines}

We train the VQ-VAE with a 5-layer GIN encoder and 1-layer MLP decoder. For NASBench-101/201, encoder hidden sizes are 128 for the first four layers and 16 for the last; for DARTS, they are 256 and 16. We use Adam with learning rate $10^{-4}$, codebook size $K=512$, latent dimension $D=16$, commitment weight $\beta=0.25$, and decay $\gamma=0.99$. Training converges within 50 epochs on NASBench-101/201, with early stopping on DARTS. The decoder-only transformer has 4 layers and 4 heads, optimized with AdamW for 100 epochs at $3\times10^{-4}$. We compare Arch-VQ with unsupervised VAE-based and unconditional generative baselines; full details and ablations are in the Appendix.

\section{Main Results}
We evaluate Arch-VQ across three NAS benchmarks on representation quality, downstream performance prediction, and latent space analysis.

\subsection{Representation Quality}

\begin{table}[t]
    \centering
    \small
     \caption{NAS-Bench-101 comparison with baseline methods. $^{*}$$^{\dagger}$ denote results from \cite{lukasik2021smooth}, \cite{lukasik2022learning}. Arch-VQ t=0.7 and t=2.4 share the same VQ-VAE and transformer but use different inference temperatures: t=0.7 optimises \textit{Absolute Uniqueness}, t=2.4 optimises \textit{Absolute Novelty}. Results in percentages.}
    \label{tab:nb-101}
    \setlength{\tabcolsep}{4pt}
    \resizebox{\linewidth}{!}{%
    \begin{tabular}{lcccccc}
    \toprule 
        Method  & \makecell{Reconstruct\\Accuracy} 
        & Validity 
        & Uniqueness 
        & \makecell{Absolute\\Uniqueness} 
        & Novelty 
        & \makecell{Absolute\\Novelty} \\
        \midrule
        D-VAE\textsuperscript{*} \protect\cite{zhang2019d}& 25.89 & 82.55 & 19.84 & 16.38 & 16.52 & 13.64 \\
        DGMG\textsuperscript{*} \protect\cite{Li2018LearningGraphs}& \textbf{99.99} & 89.70 & 29.24 & 26.23 & 16.72 & 15.00 \\
        SVGe\textsuperscript{*} \protect\cite{lukasik2021smooth}& 99.57 & 79.16 & 32.10 & 25.41 & 16.37 & 12.96 \\
        Arch2Vec \protect\cite{yan2020does}  & 98.13 & 35.36 & 99.73 & 35.27 & \textbf{85.54} & 30.26 \\
        AG-Net\textsuperscript{$\dagger$} \cite{lukasik2022learning} & - & 71.69 & 97.92 & 70.20  & 62.30 & \textbf{44.66} \\
        Arch-VQ (t=0.7)& 97.78 & \textbf{89.73} & 95.69 & \textbf{85.86}  & 11.16 & 10.01 \\
        Arch-VQ (t=2.4) & 97.78 & 42.22 & \textbf{99.81} & 42.14  & 60.76 & 25.62\\
    \bottomrule 
    \end{tabular}}
    \end{table}

Tables \ref{tab:nb-101}, \ref{tab:nb-201}, and \ref{tab:darts} compare representation quality across the three datasets against unsupervised VAE-based and unconditional generative baselines. For \textit{Arch-VQ}, and reproduced baselines, we ran experiments with five different seeds and report the mean. Appendix includes standard deviation values. All results are based on 10,000 generated architectures.

\subsubsection{NAS-Bench-101 results}
In Table~\ref{tab:nb-101}, we compare Arch-VQ against unsupervised VAE-based methods D-VAE, DGMG, SVGe, Arch2Vec~\cite{zhang2019d,Li2018LearningGraphs,lukasik2021smooth,yan2020does}, and unconditional generative method AG-Net~\cite{lukasik2022learning}. We present two variants of Arch-VQ, both sharing the same VQ-VAE and transformer, with sampling temperature varied during generation to optimise for Absolute Uniqueness and Absolute Novelty respectively.

\begin{figure}[h]
    \centering
    \includegraphics[width=1\textwidth]{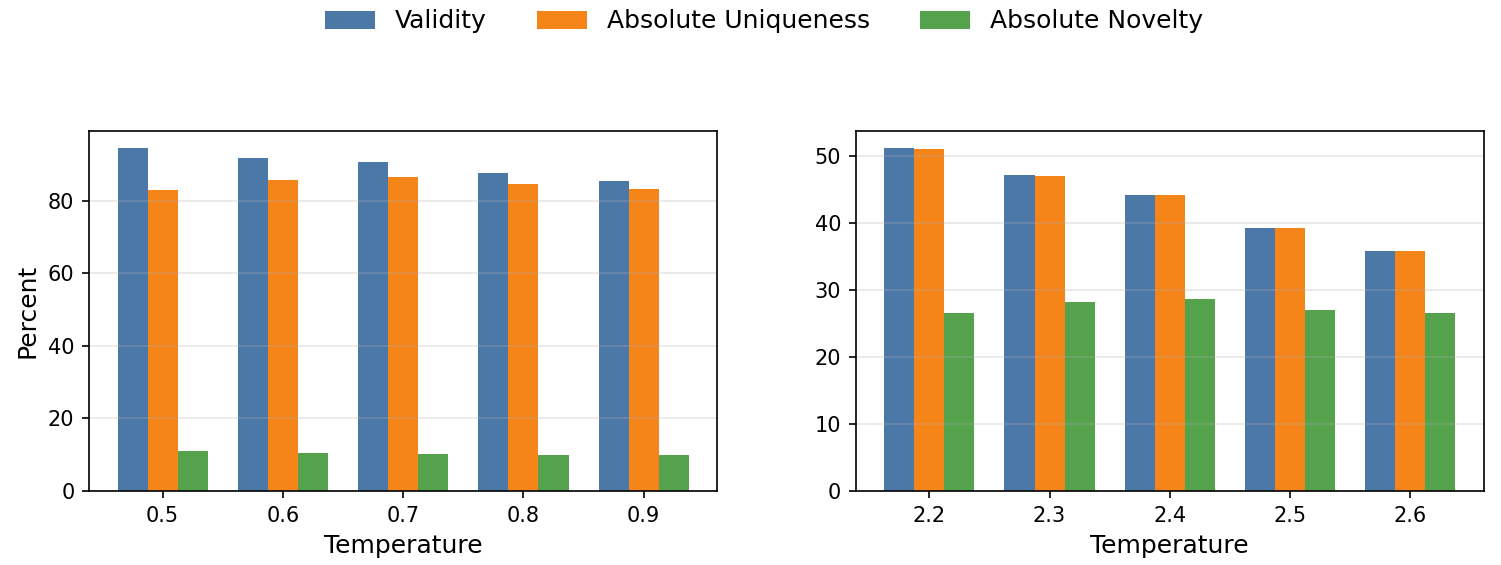}
    \caption{Validity, Absolute Uniqueness, and Absolute Novelty as a function of sampling temperature for Arch-VQ on NAS-Bench-101, optimised for Absolute Uniqueness (left) and Absolute Novelty (right). Optimal points are $t=0.7$ and $t=2.4$ respectively.}
    \label{fig:temperature_plot}
\end{figure}

Arch-VQ consistently outperforms prior VAE-based methods and AG-Net across all metrics, consistent with discrete latent modelling being better aligned with the combinatorial nature of architecture spaces. Increasing the sampling temperature improves exploration and absolute novelty at the cost of a moderate reduction in validity, enabling a controllable trade-off between sample reliability and novelty through temperature scaling, as illustrated in Figure~\ref{fig:temperature_plot}.

\subsubsection{NAS-Bench-201 results}
In Table~\ref{tab:nb-201}, we compare Arch-VQ with unsupervised VAE-based methods DGMG, SVGe, and Arch2Vec~\cite{Li2018LearningGraphs,lukasik2021smooth,yan2020does}, and unconditional generative baselines AG-Net, MetaD2A, and DiffusionNAG~\cite{lukasik2022learning,lee2021rapid,an2024diffusionnag}.

Arch-VQ achieves the best overall representation quality on NAS-Bench-201, outperforming both VAE-based and generative baselines in absolute uniqueness, consistent with its discrete latent space being better suited to the combinatorial structure of architecture search. Increasing the sampling temperature shifts generation toward exploration, achieving the best absolute novelty among all compared methods, at a moderate cost to validity. Together, the NAS-Bench-101 and NAS-Bench-201 results confirm that Arch-VQ produces higher-quality and more useful generated architectures than continuous baselines, while offering controllable diversity through temperature scaling.

\begin{table}[t]
    \centering
    \small
    \caption{NAS-Bench-201 comparison with baseline methods. $^{\ddagger}$$^{*}$$^{\dagger}$ denote results from \cite{lukasik2021smooth,lukasik2022learning,lee2021rapid}. Arch-VQ t=2.4 and t=2.9 share the same VQ-VAE and transformer but use different inference temperatures: t=2.4 optimises \textit{Absolute Uniqueness}, t=2.9 optimises \textit{Absolute Novelty}. Results in percentages.}
    \label{tab:nb-201}
    \setlength{\tabcolsep}{4pt}
    \resizebox{\linewidth}{!}{%
    \begin{tabular}{lcccccc}
    \toprule 
       Method  & \makecell{Reconstruct\\Accuracy} 
        & Validity 
        & Uniqueness 
        & \makecell{Absolute\\Uniqueness} 
        & Novelty 
        & \makecell{Absolute\\Novelty} \\
       \midrule
       DGMG\textsuperscript{*} \protect\cite{Li2018LearningGraphs}& 99.97 & \textbf{100.0} & 5.35 & 5.35 & 12.62 & 12.62\\
       SVGe\textsuperscript{*} \protect\cite{lukasik2021smooth}& \textbf{99.99} & \textbf{100.0} & 8.28 & 8.28 & 10.24 & 10.24\\
       Arch2Vec \protect\cite{yan2020does}&\textbf{99.99}  & 74.31 & 93.95 & 69.63 & 59.43 & 43.46  \\
       AG-Net\textsuperscript{$\dagger$} \cite{lukasik2022learning} & - & 99.97 & 73.61 & 73.59 & 10.03 & 10.03 \\
       MetaD2A\textsuperscript{$\ddagger$} \cite{lee2021rapid} & - & \textbf{100.0} & 35.19 & 35.19 & 67.31 & 67.31 \\
       DiffusionNAG \cite{an2024diffusionnag} & - & \textbf{100.0} &  73.70 & 73.70 & 9.88 & 9.88 \\
       Arch-VQ (t=2.4) & 99.79 & 97.52 & 95.66 & \textbf{93.30} & 69.32 & 67.65 \\
       Arch-VQ (t=2.9) & 99.79 & 89.66 & \textbf{99.15} & 88.91 & \textbf{91.02} & \textbf{81.64} \\
    \bottomrule
    \end{tabular}}
\end{table}

\begin{table}[h]
     \centering
    \small
    \caption{DARTS comparison with baseline methods. \textsuperscript{$\dagger$} denotes results from \protect\cite{lukasik2022learning}.}
    \label{tab:darts}
    \setlength{\tabcolsep}{4pt}
    \resizebox{\linewidth}{!}{%
    \begin{tabular}{lcccccc}
    \toprule 
       Method  & \makecell{Reconstruct\\Accuracy} 
        & Validity 
        & Uniqueness 
        & \makecell{Absolute\\Uniqueness} 
        & Novelty 
        & \makecell{Absolute\\Novelty} \\
       \midrule
       Arch2Vec \protect\cite{yan2020does}& \textbf{100.0} & 25.33 & \textbf{100.0} & 25.33 & \textbf{100.0} & 25.33 \\
       AG-Net\textsuperscript{$\dagger$} \protect\cite{lukasik2022learning} & - & 42.27 & \textbf{100.0} & 42.27 & \textbf{100.0} & 42.27 \\
       Arch-VQ (t=0.7) & 98.40  & \textbf{99.46} & \textbf{100.0} & \textbf{99.46} & 99.86  & \textbf{99.32} \\
    \bottomrule 
    \end{tabular}}
\end{table}

\subsubsection{DARTS results}
In Table~\ref{tab:darts}, we compare Arch-VQ against Arch2Vec and AG-Net in the DARTS search space, which is substantially larger and less constrained than NAS-Bench-101 and NAS-Bench-201, making valid generation considerably more difficult for continuous baselines.

Arch-VQ achieves near-perfect validity, absolute uniqueness, and absolute novelty, while both continuous baselines are heavily limited by low validity. These results show that the advantage of discrete autoregressive modelling becomes more pronounced as search space grows, where continuous latent sampling struggles to preserve structural constraints for valid architecture generation.

\subsection{Downstream Utility for Performance Prediction}
\label{Sec: predict_perf}

In this section, we examine whether imposing a discrete autoregressive prior on architecture representations improves the effectiveness of a downstream neural performance predictor. To test this hypothesis, we compare Arch-VQ against Arch2Vec\cite{yan2020does}, the continuous counterpart of Arch-VQ. To isolate the effect of the representation itself, we adopt a simple autoregressive predictor: a one-layer LSTM regressor with hidden size 64 and 4 heads. We choose an LSTM as it naturally consumes sequential inputs, making it equally applicable to both Arch-VQ's discrete code sequences and the concatenated node-level sequences constructed for Arch2Vec, ensuring a fair comparison. We evaluate on NAS-Bench-101 and NAS-Bench-201; for the DARTS search space, where ground-truth labels are unavailable, we instead use the NAS-Bench-301 proxy dataset \cite{akhauri2024encodings}. For Arch-VQ, we use the learned discrete embeddings directly. For each benchmark, we train the predictor on 6 different training splits and report performance on the full dataset using Kendall’s Tau, which measures the quality of ranking, which is adopted in NAS performance predictor literature \cite{akhauri2024encodings}.

\begin{figure}[t]
    \centering
    \includegraphics[width=1\textwidth]{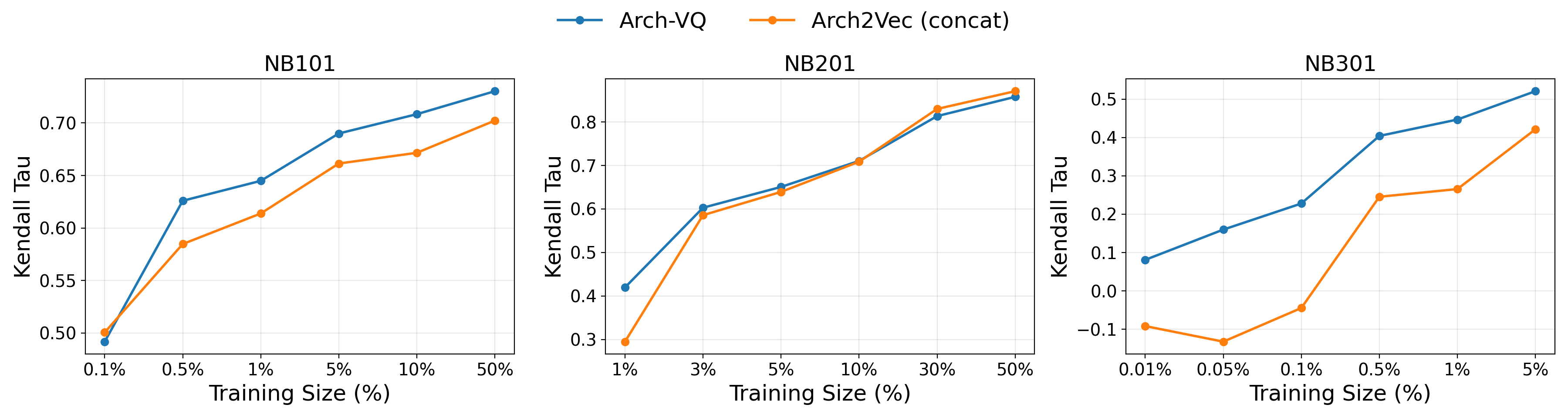}
    \caption{Illustration of neural predictor performance of Arch-VQ and Arch2Vec (concat) on NASBench-101, NASBench-201 and NASBench-301 respectively. We train using different dataset percentages and report Kendall Tau over the full dataset. Mean values over 10 seeds are plotted in the graphs.}
    \label{fig:np_kendall_tau}
\end{figure}

Figures \ref{fig:np_kendall_tau}(a–c) present the results on NAS-Bench-101, NAS-Bench-201, and NAS-Bench-301, respectively. The results show that, Arch-VQ outperforms the continuous prior on NAS-Bench-101 and NAS-Bench-301, while remaining competitive on NAS-Bench-201. These findings suggest that the discrete autoregressive prior produces representations that are more amenable to downstream prediction, likely because it captures architecture structure in a way that is better aligned with the sequential inductive bias of the predictor.

\subsection{Downstream Utility for Unconditional Generation and Ranking}

In Table \ref{tab:nb201_unconditional_sort}, we demonstrate downstream applicability in unconditional generation and sorting. We conducted an experiment with generating architectures from the latent spaces of the NASBench-201 dataset. We generate 10,000 architectures and sort 1000 architectures using the neural predictor trained in Section \ref{Sec: predict_perf}. Since DiffusionNAG\cite{an2024diffusionnag} is a conditional generation algorithm, we compare here only their unconditional generation results, similar to our setting. Unlike baseline methods, which frequently generate degenerate architectures with near-random performance (Min around 1–10\%), Arch-VQ maintains high worst-case performance, demonstrating that the learned discrete prior preserves structural validity even under unconditional generation. 

\subsection{Analyzing the Learned Discrete Latent Space}

\begin{table}[t]
    \centering
    \small
    \caption{Statistics of generated NASBench-201 architectures. Results in \textsuperscript{*} taken from \cite{an2024diffusionnag}.}
    \label{tab:nb201_unconditional_sort}
    \begin{tabular}{llccccc}
        \toprule
            &
            & \makecell{Oracle\\Top-1000\textsuperscript{*}}
            & Random\textsuperscript{*}
            & MetaD2A\textsuperscript{*}
            & \makecell{DiffusionNAG\\Uncond.\ + Sort\textsuperscript{*}}
            & \makecell{Arch-VQ\\Uncond.\ + Sort} \\
        \midrule
        \multirow{3}{*}{CIFAR10} 
            & Max  & 94.37 & \textbf{94.37} & \textbf{94.37} & \textbf{94.37} & \textbf{94.37} \\
            & Mean & 93.50 & 87.12 & 91.52 & 90.77 & \textbf{92.90} \\
            & Min  & 93.18 & 10.00 & 10.00 & 10.00 & \textbf{90.41} \\
        \midrule
        \multirow{3}{*}{CIFAR100} 
            & Max  & 73.51 & 72.74 & \textbf{73.51} & 73.16 & \textbf{73.51} \\
            & Mean & 70.62 & 61.59 & 67.14 & 66.37 & \textbf{69.47} \\
            & Min  & 69.91 & 1.00  & 1.00  & 1.00  & \textbf{63.82} \\
        \bottomrule
    \end{tabular}
\end{table} 

\subsubsection{Autoregressive property of the sequences}To evaluate whether the learned representation preserves the autoregressive structure of the latent sequence, we performed a permutation sensitivity analysis. We sampled 1,000 latent code sequences from NASBench-101 and generated 100 random permutations for each sequence.  After decoding the permuted sequences, only 5.1\% produced the same architecture as the original, while 94.9\% produced different ones, as summarized in Table \ref{tab:ar_property}. This indicates that the generation process strongly depends on the order of the latent codes, confirming that the autoregressive prior captures meaningful sequential dependencies in the architecture representation.


\begin{table}[t]
\centering
\caption{Architecture consistency under original and permuted latent code sequences.}
\setlength{\tabcolsep}{8pt}
\begin{tabular}{lcc}
\hline
Sequence setting & Decoded to original & Not decoded to original \\
\hline
Original ordering & 1000 (100\%) & 0 (0\%) \\
Permuted ordering & 51 (5.1\%) & 949 (94.9\%) \\
\hline
\label{tab:ar_property}
\end{tabular}
\end{table}

\subsubsection{Distribution matching} 
\begin{figure}[h]
    \centering
    \includegraphics[width=1\textwidth]{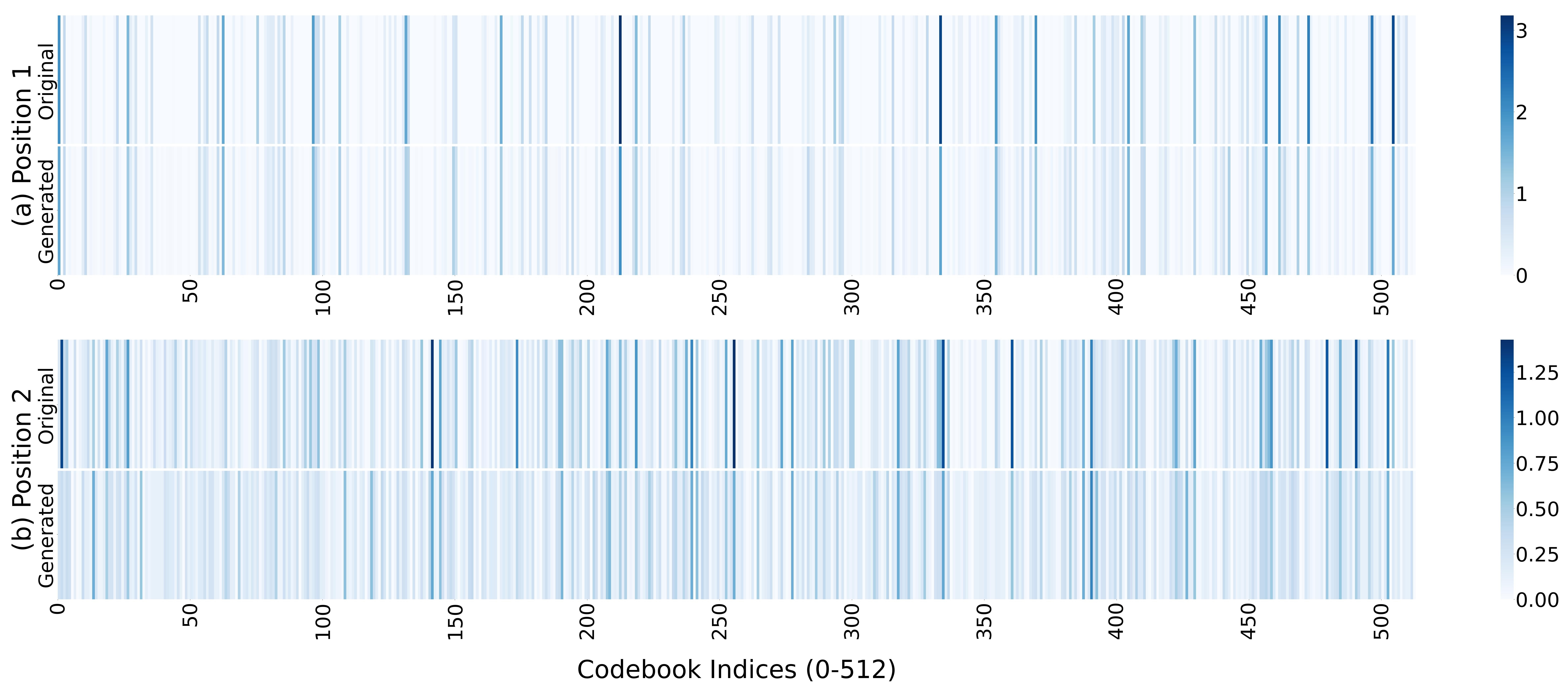}
    \caption{Original and generated novel architecture distributions for \textit{Arch-VQ}. Heatmaps show codebook-index distributions at sequence positions 1 and 2: the top row shows the training distribution, and the bottom row shows the generated novel architecture distribution.}
    \label{fig:ori_vs_nov_cb}
\end{figure}

We compare the distribution of training architecture sequences with novel sequences generated by the transformer using \textit{Arch-VQ (t=2.4)} on NASBench-101. Figure \ref{fig:ori_vs_nov_cb} shows the distributions at sequence positions 1 and 2, with results for all positions in the Appendix. The generated sequences closely match the empirical distribution across positions, indicating that the learned autoregressive prior captures the underlying architecture distribution while still producing novel samples.

\section{Ablation Studies}

\subsection{Discretization vs Autoregression}
Table \ref{tab:ablation_nb201} ablates two core design choices on NAS-Bench-201. Without discretization (VAE + Transformer), Absolute Uniqueness and Absolute Novelty drop sharply, confirming that VQ-VAE's discrete latent space is essential for diverse generation. Substituting a non-autoregressive prior over discrete codes recovers moderate uniqueness but limits exploration, underscoring the role of the autoregressive transformer. Arch-VQ combining both components achieves the best results, with temperature offering a controllable fidelity–exploration trade-off: t=2.4 favors validity, and absolute uniqueness, while t=2.9 prioritizes absolute novelty at a modest validity cost.

\begin{table}[t]
    \centering
    \small
    \caption{Ablation of VQ-VAE and autoregressive prior on NAS-Bench-201}
    \label{tab:ablation_nb201}
    \setlength{\tabcolsep}{4pt}
    \resizebox{\linewidth}{!}{%
    \begin{tabular}{lccccc}
    \toprule 
        Method 
        & Validity 
        & Uniqueness 
        & \makecell{Absolute\\Uniqueness} 
        & Novelty 
        & \makecell{Absolute\\Novelty} \\
        \midrule
        VAE + Transformer & 100.0 & 28.12 & 28.12  & 11.18 & 11.18 \\
        VQ-VAE + Standard Prior & 100.0 & 74.84 & 74.84  & 13.76 & 13.76 \\
        Arch-VQ (t=2.4) & 97.52 & 95.66 & \textbf{93.30} & 69.32 & 67.65 \\
        Arch-VQ (t=2.9) & 89.66 & \textbf{99.15} & 88.91 & \textbf{91.02} & \textbf{81.64} \\
    \bottomrule 
    \end{tabular}}
    \end{table}

\subsection{Hyperparameter Sensitivity Analysis}

\begin{figure}[h]
    \centering
    \begin{subfigure}{0.32\linewidth}
        \centering
        \includegraphics[width=\linewidth]{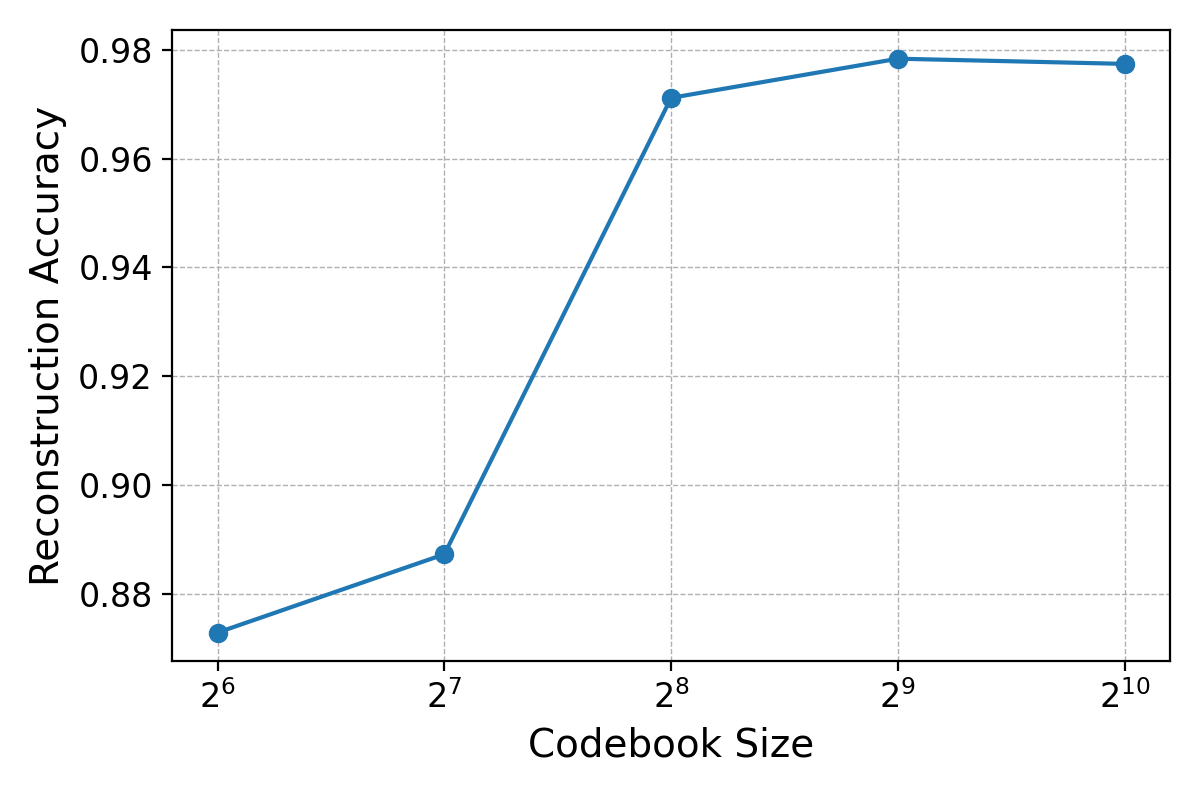}
        \caption{Codebook Size}
    \end{subfigure}
    \begin{subfigure}{0.32\linewidth}
        \centering
        \includegraphics[width=\linewidth]{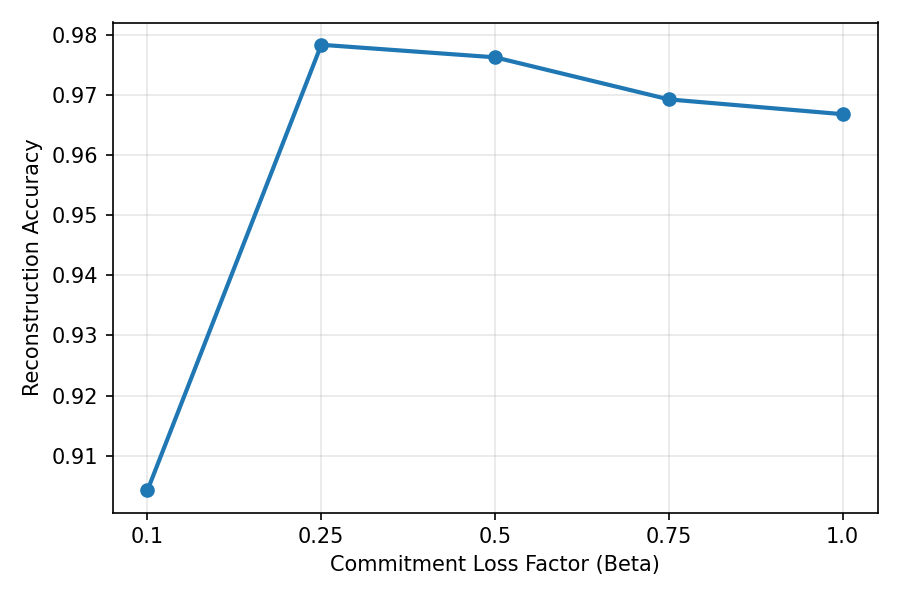}
        \caption{ $\beta$ Values}
    \end{subfigure}
    \begin{subfigure}{0.32\linewidth}
        \centering
        \includegraphics[width=\linewidth]{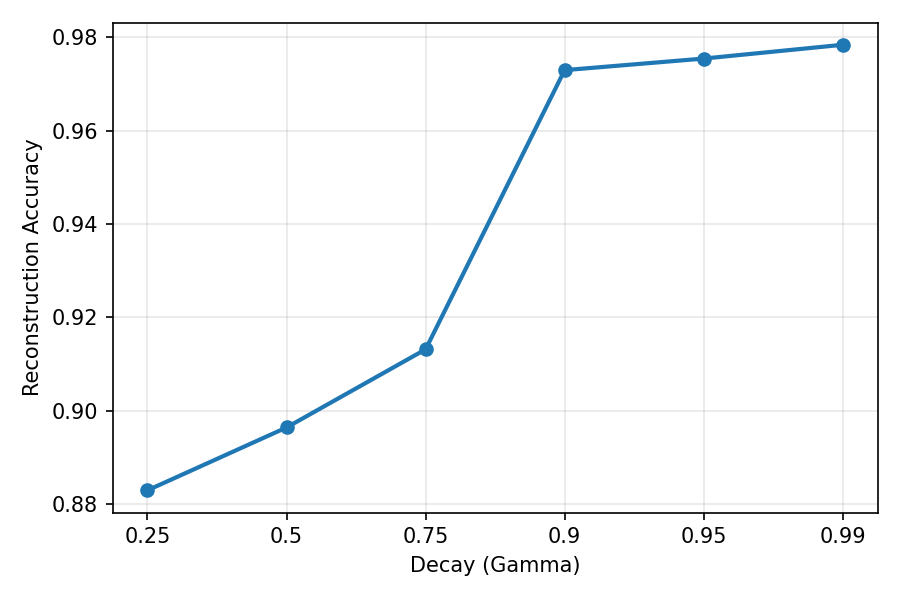}
        \caption{$\gamma$ Values}
    \end{subfigure}
    \caption{Sensitivity analysis on different hyperparameter choices of Arch-VQ}
    \label{fig:ablation_hp}
\end{figure}

In this section, we conduct an ablation study on key hyperparameters of Arch-VQ under the NASBench-101 setting. In each experiment, we vary one hyperparameter while keeping all others fixed, and report the reconstruction accuracy of the VQ-VAE. Specifically, we evaluate codebook sizes \(K \in \{128,\allowbreak 256,\allowbreak 512,\allowbreak 1024\}\), commitment weight \(\beta \in \{0.1,0.25,0.5,0.75,1.0\}\), and decay parameter gamma \(\gamma \in \{0.25,0.5,0.75,0.9,0.95,0.99\}\). Figure~\ref{fig:ablation_hp} shows that reconstruction accuracy improves as the codebook size increases up to 512, after which the gains saturate. For the commitment weight, intermediate values perform best, with \(\beta = 0.25\) achieving the highest reconstruction accuracy, suggesting that moderate regularization is important for stable codebook utilization. For $\gamma$ value, we chose 0.99 performing the best result. Overall, the results indicate that Arch-VQ is reasonably robust to moderate hyperparameter variation. Based on these findings, we use \(K = 512\), \(\beta = 0.25\), and \(\gamma = 0.99\), in all subsequent experiments.

\section{Conclusion, Limitations and Future Work}

This work revisits unsupervised neural architecture representation learning from the perspective of discrete latent modeling. We introduced Arch-VQ, a framework that combines a VQ-VAE for learning discrete architecture codes with an autoregressive Transformer prior over code sequences. Experimental results on NASBench-101, NASBench-201, and DARTS demonstrate that Arch-VQ learns higher-quality latent representations than VAE-based and other generative baselines, while remaining effective for downstream performance prediction. These results suggest that discrete latent spaces provide a more suitable inductive bias for neural architecture representation learning than continuous latent spaces.

The current framework nevertheless has several limitations. First, the current pipeline is restricted to cell-based search spaces, which may limit its applicability to broader architecture design settings. Second, during inference, using a higher sampling temperature increases randomness in the generated code sequences, which can lead to invalid outputs, such as empty sequences or sequences with incorrect lengths. In practice, this requires additional post-processing to identify and filter invalid samples. Moreover, we focus specifically on the representation learning aspect of neural architectures. Accordingly, we present an unconditional generation pipeline combined with a neural predictor, rather than a full NAS optimization algorithm. Overall, Arch-VQ highlights the potential of discrete latent modeling as a strong foundation for future research on conditional, guided, and task-aware architecture generation.

\begin{credits}
\subsubsection{Use of generative AI in the writing process}
The authors used generative AI tools for language editing and readability, and take full responsibility for the manuscript.

\end{credits}
%
%
%
\bibliographystyle{splncs04}
\bibliography{ref}

@inproceedings{akhauri2024encodings,
  title={Encodings for Prediction-based Neural Architecture Search},
  author={Akhauri, Yash and Abdelfattah, Mohamed S},
  booktitle={International Conference on Machine Learning},
  pages={740--759},
  year={2024},
  organization={PMLR}
}

@inproceedings{an2024diffusionnag,
  title={DIFFUSIONNAG: PREDICTOR-GUIDED NEURAL ARCHITECTURE GENERATION WITH DIFFUSION MODELS},
  author={An, Sohyun and Lee, Hayeon and Jo, Jaehyeong and Lee, Seanie and Hwang, Sung Ju},
  booktitle={12th International Conference on Learning Representations, ICLR 2024},
  year={2024}
}

@inproceedings{ayilo2025diffusion,
  title={Diffusion-based unsupervised audio-visual speech enhancement},
  author={Ayilo, Jean-Eudes and Sadeghi, Mostafa and Serizel, Romain and Alameda-Pineda, Xavier},
  booktitle={ICASSP 2025-2025 IEEE International Conference on Acoustics, Speech and Signal Processing (ICASSP)},
  pages={1--5},
  year={2025},
  organization={IEEE}
}

@inproceedings{devlin2019bert,
  title={Bert: Pre-training of deep bidirectional transformers for language understanding},
  author={Devlin, Jacob and Chang, Ming-Wei and Lee, Kenton and Toutanova, Kristina},
  booktitle={Proceedings of the 2019 conference of the North American chapter of the association for computational linguistics: human language technologies, volume 1 (long and short papers)},
  pages={4171--4186},
  year={2019}
}

@article{dongbench,
  title={NAS-BENCH-201: EXTENDING THE SCOPE OF RE-PRODUCIBLE NEURAL ARCHITECTURE SEARCH},
  author={Dong, Xuanyi and Yang, Yi}
}

@inproceedings{esser2021taming,
  title={Taming transformers for high-resolution image synthesis},
  author={Esser, Patrick and Rombach, Robin and Ommer, Bjorn},
  booktitle={Proceedings of the IEEE/CVF conference on computer vision and pattern recognition},
  pages={12873--12883},
  year={2021}
}

@inproceedings{he2020momentum,
  title={Momentum contrast for unsupervised visual representation learning},
  author={He, Kaiming and Fan, Haoqi and Wu, Yuxin and Xie, Saining and Girshick, Ross},
  booktitle={Proceedings of the IEEE/CVF conference on computer vision and pattern recognition},
  pages={9729--9738},
  year={2020}
}

@inproceedings{jansen2018unsupervised,
  title={Unsupervised learning of semantic audio representations},
  author={Jansen, Aren and Plakal, Manoj and Pandya, Ratheet and Ellis, Daniel PW and Hershey, Shawn and Liu, Jiayang and Moore, R Channing and Saurous, Rif A},
  booktitle={2018 IEEE international conference on acoustics, speech and signal processing (ICASSP)},
  pages={126--130},
  year={2018},
  organization={IEEE}
}

@inproceedings{lee2021rapid,
  title={Rapid Neural Architecture Search by Learning to Generate Graphs from Datasets},
  author={LEE, HAYEON and Hyung, Eunyoung and Hwang, Sung Ju},
  booktitle={The Ninth International Conference on Learning Representations},
  year={2021},
  organization={The International Conference on Learning Representations (ICLR)}
}

@inproceedings{liu2018darts,
title={{DARTS}: Differentiable Architecture Search},
author={Hanxiao Liu and Karen Simonyan and Yiming Yang},
booktitle={International Conference on Learning Representations},
year={2019}
}

@inproceedings{lukasik2021smooth,
  title={Smooth variational graph embeddings for efficient neural architecture search},
  author={Lukasik, Jovita and Friede, David and Zela, Arber and Hutter, Frank and Keuper, Margret},
  booktitle={2021 International Joint Conference on Neural Networks (IJCNN)},
  pages={1--8},
  year={2021},
  organization={IEEE}
}

@inproceedings{lukasik2022learning,
  title={Learning where to look--generative nas is surprisingly efficient},
  author={Lukasik, Jovita and Jung, Steffen and Keuper, Margret},
  booktitle={European Conference on Computer Vision},
  pages={257--273},
  year={2022},
  organization={Springer}
}

@article{luo2018neural,
  title={Neural architecture optimization},
  author={Luo, Renqian and Tian, Fei and Qin, Tao and Chen, Enhong and Liu, Tie-Yan},
  journal={Advances in neural information processing systems},
  volume={31},
  year={2018}
}

@article{mikolov2013distributed,
  title={Distributed representations of words and phrases and their compositionality},
  author={Mikolov, Tomas and Sutskever, Ilya and Chen, Kai and Corrado, Greg S and Dean, Jeff},
  journal={Advances in neural information processing systems},
  volume={26},
  year={2013}
}

@article{van2016conditional,
  title={Conditional image generation with pixelcnn decoders},
  author={Van den Oord, Aaron and Kalchbrenner, Nal and Espeholt, Lasse and Vinyals, Oriol and Graves, Alex and others},
  journal={Advances in neural information processing systems},
  volume={29},
  year={2016}
}

@article{van2017neural,
  title={Neural discrete representation learning},
  author={Van Den Oord, Aaron and Vinyals, Oriol and others},
  journal={Advances in neural information processing systems},
  volume={30},
  year={2017}
}

@inproceedings{real2019regularized,
  title={Regularized evolution for image classifier architecture search},
  author={Real, Esteban and Aggarwal, Alok and Huang, Yanping and Le, Quoc V},
  booktitle={Proceedings of the aaai conference on artificial intelligence},
  volume={33},
  number={01},
  pages={4780--4789},
  year={2019}
}

@inproceedings{real2017large,
  title={Large-scale evolution of image classifiers},
  author={Real, Esteban and Moore, Sherry and Selle, Andrew and Saxena, Saurabh and Suematsu, Yutaka Leon and Tan, Jie and Le, Quoc V and Kurakin, Alexey},
  booktitle={International conference on machine learning},
  pages={2902--2911},
  year={2017},
  organization={PMLR}
}

@article{tjandra2019vqvae,
  title={VQVAE Unsupervised Unit Discovery and Multi-scale Code2Spec Inverter for Zerospeech Challenge 2019},
  author={Tjandra, Andros and Sisman, Berrak and Zhang, Mingyang and Sakti, Sakriani and Li, Haizhou and Nakamura, Satoshi},
  year={2019}
}

@inproceedings{
xu2018how,
title={How Powerful are Graph Neural Networks?},
author={Keyulu Xu and Weihua Hu and Jure Leskovec and Stefanie Jegelka},
booktitle={International Conference on Learning Representations},
year={2019},
}

@article{yan2020does,
  title={Does unsupervised architecture representation learning help neural architecture search?},
  author={Yan, Shen and Zheng, Yu and Ao, Wei and Zeng, Xiao and Zhang, Mi},
  journal={Advances in neural information processing systems},
  volume={33},
  pages={12486--12498},
  year={2020}
}

@article{yan2021videogpt,
  title={Videogpt: Video generation using vq-vae and transformers},
  author={Yan, Wilson and Zhang, Yunzhi and Abbeel, Pieter and Srinivas, Aravind},
  journal={arXiv preprint arXiv:2104.10157},
  year={2021}
}

@inproceedings{yang2023hotnas,
  title={Hotnas: Hierarchical optimal transport for neural architecture search},
  author={Yang, Jiechao and Liu, Yong and Xu, Hongteng},
  booktitle={Proceedings of the IEEE/CVF Conference on Computer Vision and Pattern Recognition},
  pages={11990--12000},
  year={2023}
}

@inproceedings{ying2019bench,
  title={Nas-bench-101: Towards reproducible neural architecture search},
  author={Ying, Chris and Klein, Aaron and Christiansen, Eric and Real, Esteban and Murphy, Kevin and Hutter, Frank},
  booktitle={International conference on machine learning},
  pages={7105--7114},
  year={2019},
  organization={PMLR}
}

@article{zhang2019d,
  title={D-vae: A variational autoencoder for directed acyclic graphs},
  author={Zhang, Muhan and Jiang, Shali and Cui, Zhicheng and Garnett, Roman and Chen, Yixin},
  journal={Advances in neural information processing systems},
  volume={32},
  year={2019}
}

@inproceedings{zoph2017neural,
  title={Neural Architecture Search with Reinforcement Learning},
  author={Zoph, Barret and Le, Quoc},
  booktitle={International Conference on Learning Representations},
  year={2017}
}

@techreport{OpenaiImprovingPre-Training,
    title = {{Improving Language Understanding by Generative Pre-Training}},
    author = {Openai, Alec Radford and Openai, Karthik Narasimhan and Openai, Tim Salimans and Openai, Ilya Sutskever},
    url = {https://gluebenchmark.com/leaderboard}
}

@article{Li2018LearningGraphs,
    title = {{Learning Deep Generative Models of Graphs}},
    year = {2018},
    author = {Li, Yujia and Vinyals, Oriol and Dyer, Chris and Pascanu, Razvan and Battaglia, Peter},
    month = {3},
    url = {http://arxiv.org/abs/1803.03324},
    arxivId = {1803.03324}
}
%
\end{document}